\documentclass[11pt]{article}
\usepackage{acl2016}
\usepackage{times}
\usepackage{url}
\usepackage{latexsym}
\usepackage{times}
\usepackage{latexsym}
\usepackage{graphicx}
\usepackage{amsmath}
\usepackage{amsfonts}
\usepackage{comment}
\usepackage{multirow}
\usepackage{array}

\aclfinalcopy % Uncomment this line for the final submission

\title{Relation extraction from clinical texts using domain invariant convolutional neural network}

\author{Sunil Kumar Sahu$^{\Psi}$ \thanks{Part of this work was done while {\it Sunil Kumar Sahu} was doing internship at {\it Excelra Knowledge Solutions Pvt Ltd}, Hyderabad,  Telangana, India.}, Ashish Anand$^{\Psi}$, Krishnadev Oruganty$^{\clubsuit}$, Mahanandeeshwar Gattu$^{\clubsuit}$  
\\$^{\Psi}$Department of Computer Science and Engineering, IIT Guwahati, Assam, India 
\\$^{\clubsuit}$Excelra Knowledge Solutions Pvt Ltd, Hyderabad,  Telangana, India 
\mbox{}\\
{\tt \{sunil.sahu, anand.ashish\}\@iitg.ernet.in}\\
{\tt \{krishnadev.oruganty, nandu.gattu\}\@gvkbio.com}
}

\date{}

\begin{document}
\maketitle
\begin{abstract}
In recent years extracting relevant information from biomedical and clinical texts such as research articles, discharge summaries, or electronic health records have been a subject of many research efforts and shared challenges. Relation extraction is the process of detecting and classifying the semantic relation among entities in a given piece of texts. Existing models for this task in biomedical domain use either manually engineered features or kernel methods to create feature vector. These features are then fed to classifier for the prediction of the correct class. It turns out that the results of these methods are highly dependent on quality of user designed features and also suffer from curse of dimensionality. In this work we focus on extracting relations from clinical discharge summaries. Our main objective is to exploit the power of convolution neural network (CNN) to learn features automatically and thus reduce the dependency on manual feature engineering. We evaluate performance of the proposed model on i2b2-2010 clinical relation extraction challenge dataset. Our results indicate that convolution neural network can be a good model for relation exaction in clinical text without being dependent on expert's knowledge on defining quality features.
\end{abstract}

\section{Introduction}

The increasing amount of biomedical and clinical texts such as research articles, clinical trials, discharge summaries, and other texts created by social network users, represents immeasurable source of information. Automatic extraction of relevant information from these resources can be useful for many applications such as drug repositioning, medical knowledge base creation etc. 
The performance of concept entity recognition systems for detecting mention of proteins, genes, drugs, diseases, tests and treatments has achieved sufficient level of accuracy, which gives us opportunity for using these data to do next level tasks of natural language processing (NLP). Relation extraction is the process of identifying how given entities are related in considered sentence or text. As given in the example sentence [S1] below, the entities  {\it Lexix} and {\it congestive heart failure} are related by {\it treatment administered medical problem} relation. These relations are important for other upper level NLP tasks and also in biomedical and clinical research \cite{Shang2011}.

[S1]: {\it He was given {\bf Lexix} to prevent him from {\bf congestive heart failure} .}

Relation extraction task in unstructured text has been modeled in many different ways. 
{\it co-occurrence} based methods due to their simplicity and flexibility are most widely used methods in biomedical and clinical domain. In co-occurrence analysis it is assumed that if two entities are coming together in many sentences, their must be a relation between them \cite{bunescu2006,Song:2011}.  Quite obviously this method can not differentiate types of relations and suffers from low precision and recall. To improve its results, different statistical measures such as point wise mutual information, chi-square or log-likelihood ratio  has been used in this approach \cite{stapley2000}.

Rule based methods are another commonly adapted methods for relation extraction task \cite{thomas2000,park2001,leroy2003}. Rules are created by carefully observing the syntactic and semantic patterns in relation instances. {\it Bootstrapping method}~\cite{xu2008} is used to improve the performance of rule based methods. Bootstrapping uses small number of known relation pair of each relation type as a seed and use these seeds to search patterns in huge unannotated text \cite{xu2008} in iterative fashion. %\cite{brin1998,Riloff:1999,xu2008}
%Initial seed instances are search in unannotated text and make new patterns, which are used to get more instances in iterative fashion, as a results huge number of automatically generated patterns can get. 
Bootstrapping method generates lots of irrelevant patterns too, which can be controlled by {\it distantly supervised} approach. Distantly supervised method uses large knowledge base such as UMLS or Freebase as an input and extract patterns from huge corpus for all pair of relations present in knowledge base  \cite{Mintz2009,riedel2010,roller2014}. The advantage of bootstrapping and distantly supervised methods over supervised methods is  that they do not require lots of manually labeled training data which is generally very hard to get. 

{\it Feature based methods} use sentences with predefined entities to construct feature vector through feature extraction \cite{hong2005,minard2011,rink2011}. Feature extraction is mainly based on linguistic and domain knowledge. Extracted feature vectors are used to decide correct class of relation present between entities in the sentence through any classification techniques. {\it Kernel methods} are extension
of feature based methods which utilize kernel functions to exploit rich syntactic information such as parse trees \cite{zelenko2003,culotta2004,Qian2012,Zeng14}. State of the art results have been obtained by these class of methods.

However, the performance of feature and kernel based methods are highly dependent on suitable feature set selection, which is not only tedious and time consuming task but also require domain knowledge and is dependent on other NLP systems. Often such dependencies make many
existing work less reproducible simply because of absence of the full and finer details of feature extraction.
Further often these methods lead to huge number of features and may get affected from curse of dimensionality issues \cite{Bengio03,collobert11a}. Another issue faced by these methods is feature extraction will have to be adjusted according to the data source.  As discussed earlier we are having multiple but diverse information
resources such as research articles, discharge summaries, clinical trials outcome etc.
%are an alternative to these methods for explicitly extracting features \cite{zelenko2003,culotta2004,li2008,Qian2012} through kernel functions. 
While in one hand multiple sources bring more information but the other hand it makes it challenging to extract meaningful information automatically simply because of diverse nature of the data source. For example, sentences in research articles are well formed and likely to use only well accepted technical terms. But sentences in clinical discharge summaries may not be well formed sentences instead it could be fragmented sentences with lots of acronyms or terms used only locally. Similarly social media articles may use slang or terms which are not technically used. This makes it difficult 
for above discussed methods.

Motivated by these issues, this work aims to exploit recent advances in machine learning and NLP domains to reduce such dependencies and utilize
convolutional neural network to learn important features with minimal manual dependencies. Convolution neural network has shown to be a powerful model for image processing, computer vision \cite{krizhevsky2012,karpathy2014} and subsequently in natural language processing it has given state of the art results in different tasks such as sentence classification \cite{kim14,kalchbrenner2014,hu2014,sharma2016}, relation classification \cite{Zeng14,dos2015} and semantic role labeling \cite{collobert11a}.

In this paper we propose a new framework for extracting relations among {\it problem}, {\it treatment} and {\it test} in clinical discharge summaries. In particular we use data available under clinical relation extraction task organized by Informatics for Integrating Biology and the Bedside (i2b2) in 2010 as part of i2b2/VA challenge~\cite{uzuner10a}. Extracting relations in clinical texts is more challenging compared to research articles as it contains incomplete or fragmented sentences, and lots of acronyms. %can add more challenges
Current state of the art methods heavily depend on manual feature engineering and use hundreds of thousands of features \cite{minard2011,rink2011}. Our result indicates the proposed model can outperform the current state of the art models by using only a small fraction of features. However the main
observation is the features used in our model is easy to replicate and adapt as per the data source compared to the feature sets generally used in these tasks.

\begin{figure*}
\begin{center}
  \includegraphics[width=15cm, height=6cm]{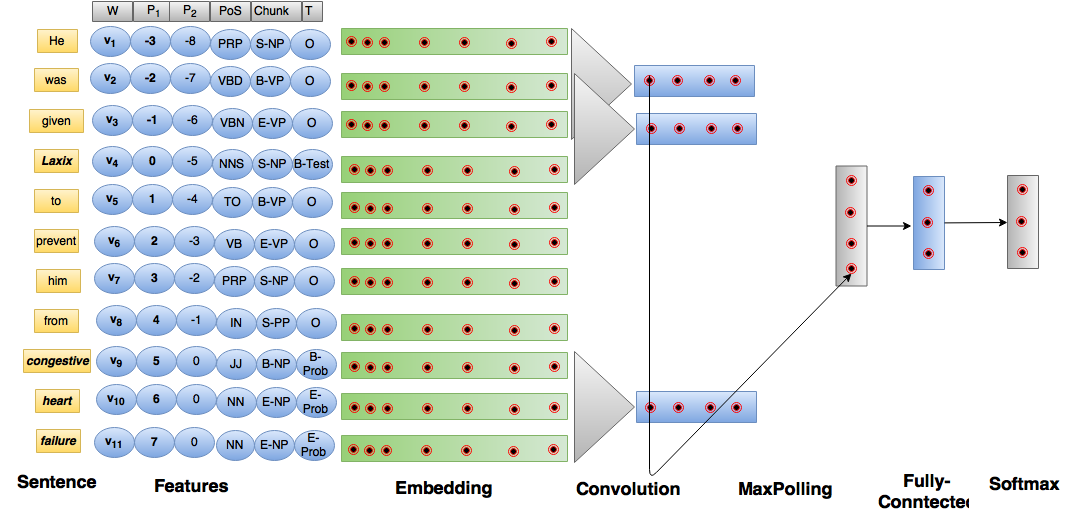}
  \caption{CNN model for relation extraction.}
  \label{fig:cnn_rel}
\end{center}
\end{figure*}

\section{Related Research}
i2b2 organized a shared task in 2010 \cite{uzuner10a}. In this challenge discharge summaries from three different sources were annotated for extracting relations among clinical entities such as {\it problem}, {\it treatment} and {\it test}. Most of the participants in this challenge used support vector machine (SVM) with manually designed features \cite{uzuner10b}. Model proposed by \newcite{rink2011} had first place in this task, which used six classes of features namely, context features, similarity features, nested related relation features, Wikipedia features, single concept features and vicinity features. They formulated the relation extraction task as a multiclass classification problem and SVM with linear kernel were used for classification. 
 
For extracting relation among disease and treatment, \newcite{Rosario04} used various graphical and neural network models. They used variety of lexical, semantic and syntactic features for classification and found that semantic features were contributing most among all. The dataset used in this study
was relatively smaller and was prepared from biomedical research articles. \newcite{li2008} proposed kernel methods for relation extraction between entities in MEDLINE\textsuperscript{\textregistered} articles. They modified the tree kernel function by incorporating trace kernel to capture richer contextual features for classifying the relation. Their results shows that tree kernel outperform other kernel methods such as word and sequence kernels for the considered task. 

Conditional random field (CRF) has been used for relation extraction between disease treatment and gene by \cite{bundschus2008}. In this experiment setting, they did not assume that entities were given, instead their model also predicted entities and its type. They developed two variants of CRF  both modeled relation extraction task as sequence labeling task.
Recently \newcite{bravo2015} proposed a system for identifying association between drug disease and target in EU-ADR dataset \cite{van2012} and named it BeeFree. BeeFree usese combination of shallow linguistic kernel and dependency kernel for identifying relations.

In contrast to above methods recently there
are few work applying convolution neural network based models \cite{Zeng14,dos2015} for {\it relation classification} in SemEval 2010 relation classification dataset \cite{hendrickx2009}. Convolution neural network used in this models are using constant length filters, and word embedding and distance embedding as features. Our model leverage on the linguistic features also and we considered {\it relation extraction} task in clinical notes which is much more informal, rich with acronyms and number of samples for each relations are not stable \cite{uzuner10a}.% \ref{sec:dataset}.
%\textwidth
\section{CNN for Clinical Relation Extraction}
The proposed model based on CNN is first summarized in the next section. Subsequent sections describe it in more detail.
\subsection{Model Architecture}
The proposed model architecture is shown in the figure \ref{fig:cnn_rel}, which takes a complete sentence with mentioned entities as an input and 
outputs a probability vector corresponding to all possible relation types.
%In input layer we extract following features for each word of the sentence
%\begin{itemize}
% \item word $w_i$
% \item relative position $P_1$ from the first mentioned entity
% \item relative position $P_2$ from the second mentioned entity
% \item PoS tag of the word $w_i$
% \item Chunk of the word $w_i$, and
% \item tag of the word $w_i$.
%\end{itemize}
Each feature is having vector representation which is initialized randomly except word embedding feature. For word embedding, we used pre-trained word
vector~\cite{muneeb15} learned on Pubmed articles using word2vec tool~\cite{mikolov13a}.
 
Embedding layer maps every feature value with its corresponding feature vectors and concatenate them. In order to get local features from each part of the sentence we have used multiple filters of different lengths \cite{kim14} in all possible continuous $n$-gram of the sentence, where $n$ is the length of filter (We have shown four filters with constant length three in the figure~\ref{fig:cnn_rel}). %, but in our experiment we learn 150 features from multiple variable length filters). 
We use max pooling over time to get global features through all filters. Here time indicates filter running over the length of the sentence. Pooled features are then fed to fully connected feed-forward neural network to make inference. In the output layer we use softmax classifier with number of outputs equal to number of possible relations between entities.

\subsection{Feature Layer}
We represent each word in the sentence with $6$ discrete features namely word itself (W), distance from the first entity ($P_1$), distance from the second entity ($P_2$), parts of speech tag of the word (PoS), chunk tag of the word (Chunk) and entity type (T). Each feature is briefly described below: %Now we are going to explain each one of this features in details:
\begin{enumerate}
    \item $W$ : Exact word appeared in the sentence.
    \item $P_1$: Distance from the first entity in terms of number of words~\cite{collobert08}. For instance in our earlier example [S1] {\it He} is at $-3$ distance and {\it prevent} is at $+2$ distance away from the first entity {\it Lexis}. This value would be zero for all words which is a part of the first entity. 
    \item $P_2$: Similar to $P_1$ but considers distance from the second entity.
    \item $PoS$: Parts of speech tag of the considered word. We use genia tagger\footnote{http://www.nactem.ac.uk/GENIA/tagger/} to obtain pos tag of each word. 
    \item $Chunk$: Chunk tag of considered word. Again we use genia tagger to obtain chunk tag of each word.
 %   \item $Dep$: Its a binary feature, it value would be yes if considered word will involve in shortest dependency path between two entities. We used Stanford parser for dependency parsing \cite{manning-EtAl:2014}
    \item $T$: Type of the considered word. For example, it would be entity type such as $B-Prob$, $I-Prob$ etc. for entity word and $Other$ for rest words following the {\it BIO} tagging convention. 
\end{enumerate}
This way a word $w \in D^1 \times D^2 \times.....D^6$, where $D^i$ is the dictionary for $i^{th}$ local features. 

\subsection{Embedding Layer}
In lookup or embedding layer each feature value is mapped to its vector representation using feature embedding matrix. Lets say $M^{i} \in \mathbb{R}^{n\times N}$ is the feature embedding matrix for $i^{th}$ local feature (here $n$ represents dimension of feature embedding and $N$ is number of possible values or size of the dictionary for $i^{th}$ local feature). Each column of $M^{i}$ is vector of corresponding value of $i^{th}$ features. Mapping can be done by taking product of one hot vector of feature value with its embedding matrix \cite{collobert08}. Suppose $a^{(i)}_j$ is the one hot vector for $j^{th}$ feature value of $i^{th}$ feature then: 

\begin{equation}
f^{(i)}_j = M^{i}\,a^{(i)}_j
\end{equation}

\begin{equation}
x^i = f^{(i)}_1 \oplus f^{(i)}_2 ....\oplus f^{(i)}_6 
\end{equation}

Here $\oplus$ is concatenation operation so $x^i \in \mathbb{R}^{(n_1+....n_6)}$ is feature vector for $i^{th}$ word in sentence and $n_k$ is dimension of $k^{th}$ feature. For word embedding we used pre-trained word vector obtained after running word2vec tool \cite{mikolov13a,mikolov13b} on huge Pubmed open source articles \cite{muneeb15}. Other feature matrix were initialized randomly at the beginning. Since number of elements in all feature dictionary except word dictionary ($D^1$) are not huge, we assume that while training these vectors will get sufficient updation.  

\subsection{Convolution Layer}
We apply convolution on text to get local features from each part of the sentence \cite{collobert08}. Consider $x^1 x^2 ..... x^m$ is the sequence of feature vectors of a sentence, where $x^i \in \mathbb{R}^d$ is a vector obtained by concatenating all feature vector of $i^{th}$ word. Let $x^{i:i+j}$ represents concatenation of  $x^i.....x^{i+j}$ feature vectors. Suppose there is a {\it filter} parameterized by weight vector $w \in \mathbb{R}^{cd}$ where $c$ is the length of filter (in figure \ref{fig:cnn_rel} filter length is three). Then output sequence of convolution layer would be
\begin{equation}
h^i = f( w \cdot x^{i:i+c-1} + b)
\end{equation}
Where $i=1, 2, \ldots m-c+1$, $.$ is dot product, $f$ is rectify linear unit (ReLu) function and $b \in \mathbb{R}$ is biased term. $w$ and $b$ are the learning parameters and will remain same for all $i=1, 2, \ldots m-c+1$.
 
\subsection{Max Pooling Layer}
Output of convolution layer length $(m-c+1)$ will vary based on number of words $m$ in the sentence. We applied max pooling \cite{collobert08} over time to get fixed length global features for whole sentence. The intuition behind using max pooling is to consider only most useful feature from entire sentence. 
\begin{equation}
z = \max_{1\leq i \leq(m-c+1)} [h^i]
\end{equation}
We have just explained the process of extracting one feature from a whole sentence using one filter. In figure \ref{fig:cnn_rel} we extracted four features using four filters of the same length three. In our experiment we use multiple such filters of variable length \cite{kim14,YinS15}. The
objective of using different length filter is to accommodate context in varying window size around words.

\subsection{Fully Connected Layer}
The output of max pooling layer is sequence $z$ came with different filters. We call this global feature because it came by taking max over entire sentence. To make classifier over extracted global feature, we used fully connected feed forward layer. Suppose $z^i \in \mathbb{R}^{l}$ is output of max pooling layer for entire filters then output of fully-connected layer would be
\begin{equation}
o^{(i)} = W^o z^i + b^o
\end{equation}
Here $W^o \in \mathbb{R}^{[r]\times l}$ and $b^o \in \mathbb{R}^{[r]}$ are parameters of neural network and $[r]$ denotes number of classes. 

\subsection{Softmax Layer}
In output layer we used softmax classifier for which objective function would be minimization of
\begin{equation}
L_i = -\log \left (  \frac{e^{o^{(i)}_{y_i}}} {\sum_{\forall j} {e^{o^{(i)}}_j }} \right)  
\end{equation}
for $i^{th}$ sentence. Here $y_i$ is correct class of relation for $i^{th}$ instance. 

\subsection{Implementation}
We experiment with filter lengths in two different experiment settings. In first, we use $100$ different filters of a fixed length in the convolutional layer, while in another set of experiments we use varying length filters, but used $100$ different filters for each varying length. So, in the first setting, we obtain $100$ features after max pooling, while in the second, we obtain $100$ times number of different length filter features. For regularization \cite{srivastava2014}, we follow \cite{kim14} and use {\it dropout} technique in output of max pooling layer. Dropout prevents co-adaptation of hidden units by randomly dropping few nodes. We set this value to $0.5$ during training and $1$ while testing. We use Adam technique \cite{adam2014} to optimize our loss function. Entire neural network parameters and feature vectors are updated while training. We have implemented the proposed model in Python language using tensorflow package \cite{tensorflow2015} and will make it available on request.  Results of each filter length were explained in results section. Dimension of word vector is set to $50$ and rest all feature embedding size is kept to $5$.

\section{Dataset and Experimental Settings}
\label{sec:dataset}
In recent years several challenges have been organized to automatically extract information from clinical texts  \cite{uzuner07,uzuner08,uzuner10a,uzuner10b,sun2013}. i2b2 has released dataset for clinical concept extraction, assertion classification and relation extraction as a part of i2b2-2010 shared task challenge. This dataset was collected from three different hospitals and was manually annotated by medical practitioners for identifying problems, treatments and test entities, and eight relation types among them. These relations were: {\it treatment caused medical problems} ({\bf TrCP}), {\it treatment administered medical problem} ({\bf TrAP}), {\it treatment worsen medical problem} ({\bf TrWP}), {\it treatment improve or cure medical problem} ({\bf TrIP}), {\it treatment was not administered because of medical problem} ({\bf TrNAP}), {\it test reveal medical problem} ({\bf TeRP}), {\it Test conducted to investigate medical problem} ({\bf TeCP}), {\it Medical problem indicates medical problems} ({\bf PIP}). \cite{uzuner10a} has given the exact definition of each relation type.

\begin{table}[ht]
\centering
\begin{tabular}{|c|c|}
\hline
{\bf Name} & {\bf Number instances} \\ \hline
%{\it TrWP}  & 132  \\ \hline
%{\it TrNAP} & 173  \\ \hline
%{\it TrIP}  & 202  \\ \hline
{\it TeCP}  & 503  \\ \hline
{\it TrCP}  & 525  \\ \hline
{\it PIP} 	& 2202 \\ \hline
{\it TrAP}  & 2616  \\ \hline
{\it TeRP}  & 3052  \\ \hline
No Relation & 55600\\ \hline
\end{tabular}
\caption{Relation types and number of instances of i2b2 dataset (partial)}
\label{tab:i2b2_stats}
\end{table}

%Original Data: 394 training reports, 477 test reports,
%Our Data: Training (73+97), Test: 256

While during the challenge original dataset had $394$ documents for training and $477$ documents for testing but when we downloaded this dataset from i2b2 website we got only $170$ documents for training and $256$ documents for testing. After preliminary experiment we found that we did not have enough training samples for all relation classes present in the dataset, therefore we decided to remove 3 relation classes along with their instances ({\it TrWP} (132 instances), {\it TrIP} (202 instances) and {\it TrNAP} (173 instances)). Statistics of the dataset is shown in the Table \ref{tab:i2b2_stats}.

For extracting relations among entities we considered all sentences having more than one entities in each discharge summary to check whether any relation exists between them or not. In our experiment we assume that entities and their types are already known like other existing works \cite{rink2011,Minard588,minard2011}. We created data sample for every pair of entities present in the sentence and labeled it with the existing relation type. For example in sentence [S2] (all continuous bold phrases are entities) entity pairs ({\it ``her white count", ``elevated"}) label would be {\it ``TeRP"}, for entity pair ({\it ``her g-csf", ``elevated"}) label would be {\it ``TrNAP"} and for ({\it ``her white count", ``her G-CSF"}) label would be "None".

[S2]: {\it {\bf Her white count} remained {\bf elevated} despite discontinuing {\bf her G-CSF} .}

\section{Results and Discussion}

\subsection{Influence of filter lengths}
\label{sec:inf_filter}
We combined the training and testing data and performed five-fold cross-validation on the available limited i2b2 dataset for all our evaluations. First we evaluate the influence of filter lengths. We experiment with selection of filter length using all features. Results as average of five-fold experiment are shown in the Table \ref{tab:i2b2_res}.

\begin{table}[ht]
\centering
\begin{tabular}{|c|c|c|c|c|}
\hline
{\bf Filter length}&{\bf Precision}&{\bf Recall}&{\bf F Score}\\ \hline
\hline
[3]         & 74.54  & 64.29 & 68.44 \\ \hline
[4]			& 74.90 & 65.50 & 69.19		\\ \hline
[5]         & 76.17 & 64.68 & 69.61 	\\ \hline
[6]			& 76.05 & 66.56 & 70.43 	\\ \hline
[7] 		& 76.76 & 64.49 & 69.23 	\\ \hline
[3,4]		& 74.96 & 64.65 & 68.91		\\ \hline
[3,5]		& 74.66 & 66.81 & 70.10		\\ \hline
[4,5]		& 74.90 & 68.20 & 70.91 	\\ \hline
{\bf [4,6]} & {\bf 76.34} & {\bf 67.35} & {\bf 71.16} 	\\ \hline
[5,6]		& 76.08 & 65.31 & 69.77		\\ \hline
[3,4,5]		& 75.83 & 65.10 & 69.30		\\ \hline
[4,5,6]		& 76.12 & 65.68 & 70.15		\\ \hline
[2,3,4,5] 	& 74.99 & 65.19 & 69.34		\\ \hline
[3,4,5,6] 	& 75.88 & 65.98 & 70.13		\\ \hline	
\end{tabular}
\caption{Comparative performance of the proposed model using filters of different lengths separately and together. Each of the models used all
features (WV+$P_1$+$P_2$+PoS+Chunk+Type) and 100 filters for each filter length.}
\label{tab:i2b2_res}
\end{table}

In case of single filter, the results indicate increasing the size of filter length generally tends to improve the performance. Using only single filter the best performance with F1 score as $70.43\%$ was obtained by using filter length of $6$. However further increasing the filter length did not improve the result. Intuitively it also seems that selection of either of too small or too large filter length may not be a good option. Filter length gives the window size to capture
context features. One can expect that too small filter length (window size) may not capture enough good context feature and too big filter length may include noise or irrelevant contexts. 

Further, we used multiple filters to see whether it improves the result. Results indicate that combination of small and mid-length filter size is perhaps the better choice. For example, combination of filter lengths $3$ and $4$ together did not improve the performance compared to the single filter length of $3$ or $4$. On the other hand combination of filter lengths $3$ and $5$, and $4$ and $5$ improved the performance compared to use of single filters of either length. It can be seen, the best result with F1 score as $71.16 \%$ is obtained by using filter lengths of $4$ and $6$ together. But adding more than two filters did not lead to performance improvement.

\subsection{Classwise Performance}
We took the best combination of filter lengths and looked at the classwise performance. Results are described in the Table~\ref{tab:clswise}. 
\begin{table}[ht]
\centering
\begin{tabular}{|c|c|c|c|c|}
\hline
{\bf Name}  &{\bf Precision}&{\bf Recall}&{\bf F Score} \\ \hline
\hline
{\it TeCP }  & 63.48 & 43.67 & 50.56 \\ \hline
{\it TrCP}   & 63.60 & 43.67 & 56.44 \\ \hline
{\it PIP}    & 67.32 & 63.30 & 64.92 \\ \hline
{\it TrAP}   & 73.49 & 65.83 & 69.23 \\ \hline
{\it TeRP}   & 82.74 & 79.88 & 81.25 \\ \hline
\end{tabular}
\caption{Class wise performance with all features (filter size : [4,6] each with 100 filters) }
\label{tab:clswise}
\end{table} 

We see from the results that as number of training examples (see Table~\ref{tab:i2b2_stats}) increases, performance of the model also improves. 
The relation class {\it TeRP} has the maximum number of training examples and the model obtained quite a good F1 score. On the other hand, the model could not perhaps able to learn well for the relation classes {\it TeCP} 
and {\it TrCP} having relatively lesser number of training examples. 

\subsection{Contribution of Each Features}
In order to investigate the contribution of each feature in final result we gradually include one feature in our model and compared the performance. Table \ref{tab:feat_contri} shows the obtained results. First we use only random vector (RV) representation along with entity types (T) (first row in the table) as a baseline for our comparison. Adding position features (2nd row) lead to approximately $15\%$ increase in recall, $7\%$ in precision and $11.7\%$ in F1 score. However including PoS and Chunk features although improved recall and F1 score by $4.3\%$ and $1.3\%$ but precision was decreased by $3.6\%$. In the second set of experiments, we first use pre-trained word vectors along with entity types (4th row) and later repeated the similar experiments as previously. Here again, inclusion of position features improved the recall by more than $14\%$ and F1 score by around $11\%$. This clearly indicates
word position relative to the entities of interest plays important role in deciding their influence in the context. Further including PoS and Chunk features also led to performance improvement. 

%If we compare this baseline with the feature set of word-embedding with entity types (4th row), we observe that Here we can observe word embedding and position embedding are important feature for considered task. $PoS$ and $Chunk$ tag will farther improve the results.
 
\begin{table}[ht]
\centering
\begin{tabular}{|c|c|c|c|c|}
\hline
{\bf Name}              &{\bf P}&{\bf R}&{\bf F}\\ \hline
\hline
{\it RV + T}            & 67.21 & 52.97   & 57.87 \\ \hline
{\it +($P_1$+$P_2$) }   & 71.86 & 60.69   & 64.66 \\ \hline
{\it +(PoS+Chunk) }     & 69.25 & 63.34   & 65.52 \\ \hline
\hline    
{\it WV + T }           & 70.75 & 59.17 & 63.82  \\ \hline
{\it +($P_1$+$P_2$)}    & 75.54 & 67.69 & 70.97  \\ \hline   
{\it +(PoS+Chunk)}      & {\bf 76.34} & {\bf 67.35} & {\bf 71.16}\\ \hline
\end{tabular}
\caption{Contribution of each features (filter size : [4,6] each with 100 filters) }
\label{tab:feat_contri}
\end{table}

\subsection{Comparison with Feature Based Method}
We could not compare our results directly with the state of the art results obtained on the i2b2 dataset as we did not have the complete dataset. We build a linear SVM classifier using similar features as defined in earlier studies \cite{rink2011} as a baseline for comparison. The following features are used for each entity pair instance: 
\begin{itemize}
\item Any word between relation arguments 
\item Any $PoS$ tags between relation arguments. We used genia tagger for $PoS$
\item Any bigram between relation arguments
\item Word preceding first and second argument
\item Any three words succeeding the first and second arguments
\item Sequence of $chunk$ tags between relation arguments. We used genia tagger for $chunk$ tag
\item String of words between relation arguments
\item First and second argument type (problem, treatment and test)
\item Order of argument type appeared in sentence
\item Distance between two arguments in terms of number of words
\item Presence of only punctuation sign between arguments.
\end{itemize}
This way we prepared attribute-value and numerical features for each instances. Table \ref{tab:comp} shows the comparison of best results obtained by the proposed model and SVM based model. Linear SVM classifier with different cost parameter $C$ was implemented using scikit learn \cite{scikit-learn}. Here again results shown are average over the 5-folds.
\begin{table}[ht]
\centering
\begin{tabular}{|c|c|c|c|c|}
\hline
{\bf Name}&{\bf P}&{\bf R}&{\bf F}\\ \hline
\hline
CNN (FL=[4,6])     			& {\bf 76.34} & {\bf 67.35} & {\bf 71.16}\\ \hline
SVM (Linear, C=0.01)		& 72.23 & 57.75 & 58.96	\\ \hline
SVM (Linear, C=0.1) 		& 73.75 & 64.18 & 67.35  \\ \hline
SVM (Linear, C=1)		    & 73.17 & 64.18 & 67.32 \\ \hline
\end{tabular}
\caption{Comparative performance of SVM and CNN with filter length [4,6] each with 100 filters}
\label{tab:comp}
\end{table}

Based on the results, We can make following observations:
\begin{itemize}
 \item Instead of SVM, other classifier could have been also used. We decided to use SVM as SVM based model with similar features obtained the best performance in the 2010 challenge.
 \item In any case we still would have to define huge number of features and only few of them would have non-zero values in any given sample or instance.
 \item The proposed model with limited number of features ($75$ * number of words in the sentence; 5 dimensional vector for 5 features other than word embedding, which is 50 dimensional vector) still gave the better performance.
 \item Consistent with our observations in the section~\ref{sec:inf_filter}, too many features trying to capture more contexts adversely affect the performance of classifier. If we
 look at the features defined above it includes features which try to capture context of all possible window size between the mentioned entities.
\end{itemize}

\section{Conclusion}
In this work we present a new framework based on CNN for extracting relations among clinical entities in clinical texts. The proposed model has shown better performance by using only a small fraction of features compared to the SVM based baseline model. Our results indicate that CNN is able to learn global features which can capture contextual features quite well and thus helps in improving the performance.

\section*{Acknowledgments}
We would like to thank i2b2 National Center for Biomedical Computing
funded by U54LM008748, for providing the clinical records originally prepared
for the Shared Tasks for Challenges in NLP for Clinical Data organized by Dr. Ozlem Uzuner, i2b2 and SUNY.

\bibliography{acl2016}

\begin{thebibliography}{}

\bibitem[\protect\citename{Abadi \bgroup et al.\egroup }2015]{tensorflow2015}
Mart{\i}n Abadi, Ashish Agarwal, Paul Barham, Eugene Brevdo, Zhifeng Chen,
  Craig Citro, Greg~S Corrado, Andy Davis, Jeffrey Dean, Matthieu Devin, et~al.
\newblock 2015.
\newblock Tensorflow: Large-scale machine learning on heterogeneous systems,
  2015.
\newblock {\em Software available from tensorflow.org}.

\bibitem[\protect\citename{Bengio \bgroup et al.\egroup }2003]{Bengio03}
Yoshua Bengio, R{\'e}jean Ducharme, Pascal Vincent, and Christian Janvin.
\newblock 2003.
\newblock A neural probabilistic language model.
\newblock {\em J. Mach. Learn. Res.}, 3:1137--1155, March.

\bibitem[\protect\citename{Bravo \bgroup et al.\egroup }2015]{bravo2015}
{\`A}lex Bravo, Janet Pi{\~n}ero, N{\'u}ria Queralt-Rosinach, Michael
  Rautschka, and Laura~I Furlong.
\newblock 2015.
\newblock Extraction of relations between genes and diseases from text and
  large-scale data analysis: implications for translational research.
\newblock {\em BMC bioinformatics}, 16(1):1.

\bibitem[\protect\citename{Bundschus \bgroup et al.\egroup
  }2008]{bundschus2008}
Markus Bundschus, Mathaeus Dejori, Martin Stetter, Volker Tresp, and Hans-Peter
  Kriegel.
\newblock 2008.
\newblock Extraction of semantic biomedical relations from text using
  conditional random fields.
\newblock {\em BMC bioinformatics}, 9(1):1.

\bibitem[\protect\citename{Bunescu \bgroup et al.\egroup }2006]{bunescu2006}
Razvan Bunescu, Raymond Mooney, Arun Ramani, and Edward Marcotte.
\newblock 2006.
\newblock Integrating co-occurrence statistics with information extraction for
  robust retrieval of protein interactions from medline.
\newblock In {\em Proceedings of the Workshop on Linking Natural Language
  Processing and Biology: Towards Deeper Biological Literature Analysis}, pages
  49--56. Association for Computational Linguistics.

\bibitem[\protect\citename{Collobert and Weston}2008]{collobert08}
Ronan Collobert and Jason Weston.
\newblock 2008.
\newblock A unified architecture for natural language processing: Deep neural
  networks with multitask learning.
\newblock In {\em Proceedings of the 25th international conference on Machine
  learning}, pages 160--167. ACM.

\bibitem[\protect\citename{Collobert \bgroup et al.\egroup }2011]{collobert11a}
Ronan Collobert, Jason Weston, L{\'e}on Bottou, Michael Karlen, Koray
  Kavukcuoglu, and Pavel Kuksa.
\newblock 2011.
\newblock Natural language processing (almost) from scratch.
\newblock {\em J. Mach. Learn. Res.}, 12:2493--2537, November.

\bibitem[\protect\citename{Culotta and Sorensen}2004]{culotta2004}
Aron Culotta and Jeffrey Sorensen.
\newblock 2004.
\newblock Dependency tree kernels for relation extraction.
\newblock In {\em Proceedings of the 42nd Annual Meeting on Association for
  Computational Linguistics}, page 423. Association for Computational
  Linguistics.

\bibitem[\protect\citename{dos Santos \bgroup et al.\egroup }2015]{dos2015}
C{\i}cero~Nogueira dos Santos, Bing Xiang, and Bowen Zhou.
\newblock 2015.
\newblock Classifying relations by ranking with convolutional neural networks.
\newblock In {\em Proceedings of the 53rd Annual Meeting of the Association for
  Computational Linguistics and the 7th International Joint Conference on
  Natural Language Processing}, volume~1, pages 626--634.

\bibitem[\protect\citename{Hendrickx \bgroup et al.\egroup
  }2009]{hendrickx2009}
Iris Hendrickx, Su~Nam Kim, Zornitsa Kozareva, Preslav Nakov, Diarmuid
  {\'O}~S{\'e}aghdha, Sebastian Pad{\'o}, Marco Pennacchiotti, Lorenza Romano,
  and Stan Szpakowicz.
\newblock 2009.
\newblock Semeval-2010 task 8: Multi-way classification of semantic relations
  between pairs of nominals.
\newblock In {\em Proceedings of the Workshop on Semantic Evaluations: Recent
  Achievements and Future Directions}, pages 94--99. Association for
  Computational Linguistics.

\bibitem[\protect\citename{Hong}2005]{hong2005}
Gumwon Hong.
\newblock 2005.
\newblock Relation extraction using support vector machine.
\newblock In {\em Natural Language Processing--IJCNLP 2005}, pages 366--377.
  Springer.

\bibitem[\protect\citename{Hu \bgroup et al.\egroup }2014]{hu2014}
Baotian Hu, Zhengdong Lu, Hang Li, and Qingcai Chen.
\newblock 2014.
\newblock Convolutional neural network architectures for matching natural
  language sentences.
\newblock In {\em Advances in Neural Information Processing Systems}, pages
  2042--2050.

\bibitem[\protect\citename{Kalchbrenner \bgroup et al.\egroup
  }2014]{kalchbrenner2014}
Nal Kalchbrenner, Edward Grefenstette, and Phil Blunsom.
\newblock 2014.
\newblock A convolutional neural network for modelling sentences.
\newblock {\em arXiv preprint arXiv:1404.2188}.

\bibitem[\protect\citename{Karpathy and Fei-Fei}2014]{karpathy2014}
Andrej Karpathy and Li~Fei-Fei.
\newblock 2014.
\newblock Deep visual-semantic alignments for generating image descriptions.
\newblock {\em arXiv preprint arXiv:1412.2306}.

\bibitem[\protect\citename{Kim}2014]{kim14}
Yoon Kim.
\newblock 2014.
\newblock Convolutional neural networks for sentence classification.
\newblock In {\em Proceedings of the 2014 Conference on Empirical Methods in
  Natural Language Processing (EMNLP)}, pages 1746--1751, Doha, Qatar, October.
  Association for Computational Linguistics.

\bibitem[\protect\citename{Kingma and Ba}2014]{adam2014}
Diederik Kingma and Jimmy Ba.
\newblock 2014.
\newblock Adam: A method for stochastic optimization.
\newblock {\em arXiv preprint arXiv:1412.6980}.

\bibitem[\protect\citename{Krizhevsky \bgroup et al.\egroup
  }2012]{krizhevsky2012}
Alex Krizhevsky, Ilya Sutskever, and Geoffrey~E Hinton.
\newblock 2012.
\newblock Imagenet classification with deep convolutional neural networks.
\newblock pages 1097--1105.

\bibitem[\protect\citename{Leroy \bgroup et al.\egroup }2003]{leroy2003}
Gondy Leroy, Hsinchun Chen, and Jesse~D Martinez.
\newblock 2003.
\newblock A shallow parser based on closed-class words to capture relations in
  biomedical text.
\newblock {\em Journal of biomedical Informatics}, 36(3):145--158.

\bibitem[\protect\citename{Li \bgroup et al.\egroup }2008]{li2008}
Jiexun Li, Zhu Zhang, Xin Li, and Hsinchun Chen.
\newblock 2008.
\newblock Kernel-based learning for biomedical relation extraction.
\newblock {\em Journal of the American Society for Information Science and
  Technology}, 59(5):756--769.

\bibitem[\protect\citename{Mikolov \bgroup et al.\egroup }2013a]{mikolov13b}
Tomas Mikolov, Kai Chen, Greg Corrado, and Jeffrey Dean.
\newblock 2013a.
\newblock Efficient estimation of word representations in vector space.
\newblock {\em arXiv preprint arXiv:1301.3781}.

\bibitem[\protect\citename{Mikolov \bgroup et al.\egroup }2013b]{mikolov13a}
Tomas Mikolov, Ilya Sutskever, Kai Chen, Greg~S Corrado, and Jeff Dean.
\newblock 2013b.
\newblock Distributed representations of words and phrases and their
  compositionality.
\newblock In {\em Advances in Neural Information Processing Systems}, pages
  3111--3119.

\bibitem[\protect\citename{Minard \bgroup et al.\egroup }2011a]{Minard588}
Anne-Lyse Minard, Anne-Laure Ligozat, Asma Ben~Abacha, Delphine Bernhard, Bruno
  Cartoni, Louise Del{\'e}ger, Brigitte Grau, Sophie Rosset, Pierre
  Zweigenbaum, and Cyril Grouin.
\newblock 2011a.
\newblock Hybrid methods for improving information access in clinical
  documents: concept, assertion, and relation identification.
\newblock {\em Journal of the American Medical Informatics Association},
  18(5):588--593.

\bibitem[\protect\citename{Minard \bgroup et al.\egroup }2011b]{minard2011}
Anne-Lyse Minard, Anne-Laure Ligozat, and Brigitte Grau.
\newblock 2011b.
\newblock Multi-class svm for relation extraction from clinical reports.
\newblock In {\em RANLP}, pages 604--609.

\bibitem[\protect\citename{Mintz \bgroup et al.\egroup }2009]{Mintz2009}
Mike Mintz, Steven Bills, Rion Snow, and Dan Jurafsky.
\newblock 2009.
\newblock Distant supervision for relation extraction without labeled data.
\newblock In {\em Proceedings of the Joint Conference of the 47th Annual
  Meeting of the ACL and the 4th International Joint Conference on Natural
  Language Processing of the AFNLP: Volume 2 - Volume 2}, ACL '09, pages
  1003--1011, Stroudsburg, PA, USA. Association for Computational Linguistics.

\bibitem[\protect\citename{Park \bgroup et al.\egroup }2001]{park2001}
Jong~C Park, Hyun~Sook Kim, and Jung-Jae Kim.
\newblock 2001.
\newblock Bidirectional incremental parsing for automatic pathway
  identification with combinatory categorial grammar.
\newblock In {\em Pacific Symposium on Biocomputing}, volume~6, pages 396--407.

\bibitem[\protect\citename{Pedregosa \bgroup et al.\egroup }2011]{scikit-learn}
F.~Pedregosa, G.~Varoquaux, A.~Gramfort, V.~Michel, B.~Thirion, O.~Grisel,
  M.~Blondel, P.~Prettenhofer, R.~Weiss, V.~Dubourg, J.~Vanderplas, A.~Passos,
  D.~Cournapeau, M.~Brucher, M.~Perrot, and E.~Duchesnay.
\newblock 2011.
\newblock Scikit-learn: Machine learning in {P}ython.
\newblock {\em Journal of Machine Learning Research}, 12:2825--2830.

\bibitem[\protect\citename{Qian and Zhou}2012]{Qian2012}
Longhua Qian and Guodong Zhou.
\newblock 2012.
\newblock Tree kernel-based protein–protein interaction extraction from
  biomedical literature.
\newblock {\em Journal of Biomedical Informatics}, 45(3):535 -- 543.

\bibitem[\protect\citename{Riedel \bgroup et al.\egroup }2010]{riedel2010}
Sebastian Riedel, Limin Yao, and Andrew McCallum.
\newblock 2010.
\newblock Modeling relations and their mentions without labeled text.
\newblock In {\em Machine Learning and Knowledge Discovery in Databases}, pages
  148--163. Springer.

\bibitem[\protect\citename{Rink \bgroup et al.\egroup }2011]{rink2011}
Bryan Rink, Sanda Harabagiu, and Kirk Roberts.
\newblock 2011.
\newblock Automatic extraction of relations between medical concepts in
  clinical texts.
\newblock {\em Journal of the American Medical Informatics Association},
  18(5):594--600.

\bibitem[\protect\citename{Roller and Stevenson}2014]{roller2014}
Roland Roller and Mark Stevenson.
\newblock 2014.
\newblock Applying umls for distantly supervised relation detection.
\newblock In {\em Proceedings of the 5th International Workshop on Health Text
  Mining and Information Analysis (Louhi)}, pages 80--84, Gothenburg, Sweden,
  April. Association for Computational Linguistics.

\bibitem[\protect\citename{Rosario and Hearst}2004]{Rosario04}
Barbara Rosario and Marti~A. Hearst.
\newblock 2004.
\newblock Classifying semantic relations in bioscience texts.
\newblock In {\em Proceedings of the 42Nd Annual Meeting on Association for
  Computational Linguistics}, ACL '04, Stroudsburg, PA, USA. Association for
  Computational Linguistics.

\bibitem[\protect\citename{Shang \bgroup et al.\egroup }2011]{Shang2011}
Yue Shang, Yanpeng Li, Hongfei Lin, and Zhihao Yang.
\newblock 2011.
\newblock Enhancing biomedical text summarization using semantic relation
  extraction.
\newblock {\em PLoS ONE}, 6(8):1--10, 08.

\bibitem[\protect\citename{Sharma \bgroup et al.\egroup }2016]{sharma2016}
Ranti~D Sharma, Samarth Tripathi, Sunil~K Sahu, Sudhanshu Mittal, and Ashish
  Anand.
\newblock 2016.
\newblock Predicting online doctor ratings from user reviews using
  convolutional neural networks.
\newblock {\em International Journal of Machine Learning and Computing},
  6(2):149.

\bibitem[\protect\citename{Song \bgroup et al.\egroup }2011]{Song:2011}
Qiang Song, Yousuke Watanabe, and Haruo Yokota.
\newblock 2011.
\newblock Relationship extraction methods based on co-occurrence in web pages
  and files.
\newblock In {\em Proceedings of the 13th International Conference on
  Information Integration and Web-based Applications and Services}, iiWAS '11,
  pages 82--89, New York, NY, USA. ACM.

\bibitem[\protect\citename{Srivastava \bgroup et al.\egroup
  }2014]{srivastava2014}
Nitish Srivastava, Geoffrey Hinton, Alex Krizhevsky, Ilya Sutskever, and Ruslan
  Salakhutdinov.
\newblock 2014.
\newblock Dropout: A simple way to prevent neural networks from overfitting.
\newblock {\em The Journal of Machine Learning Research}, 15(1):1929--1958.

\bibitem[\protect\citename{Stapley and Benoit}2000]{stapley2000}
Benjamin~J Stapley and Gerry Benoit.
\newblock 2000.
\newblock Biobibliometrics: information retrieval and visualization from
  co-occurrences of gene names in medline abstracts.
\newblock In {\em Pac Symp Biocomput}, volume~5, pages 529--540.

\bibitem[\protect\citename{Sun \bgroup et al.\egroup }2013]{sun2013}
Weiyi Sun, Anna Rumshisky, and Ozlem Uzuner.
\newblock 2013.
\newblock Evaluating temporal relations in clinical text: 2012 i2b2 challenge.
\newblock {\em Journal of the American Medical Informatics Association}, 20(5).

\bibitem[\protect\citename{TH \bgroup et al.\egroup }2015]{muneeb15}
MUNEEB TH, Sunil Sahu, and Ashish Anand.
\newblock 2015.
\newblock Evaluating distributed word representations for capturing semantics
  of biomedical concepts.
\newblock In {\em Proceedings of BioNLP 15}, pages 158--163, Beijing, China,
  July. Association for Computational Linguistics.

\bibitem[\protect\citename{Thomas \bgroup et al.\egroup }2000]{thomas2000}
James Thomas, David Milward, Christos Ouzounis, Stephen Pulman, and Mark
  Carroll.
\newblock 2000.
\newblock Automatic extraction of protein interactions from scientific.
\newblock In {\em Pacific symposium on biocomputing}, volume~5, pages 538--549.

\bibitem[\protect\citename{Uzuner \bgroup et al.\egroup }2007]{uzuner07}
{\"O}zlem Uzuner, Yuan Luo, and Peter Szolovits.
\newblock 2007.
\newblock Evaluating the state-of-the-art in automatic de-identification.
\newblock {\em Journal of the American Medical Informatics Association},
  14(5):550--563.

\bibitem[\protect\citename{Uzuner \bgroup et al.\egroup }2008]{uzuner08}
{\"O}zlem Uzuner, Ira Goldstein, Yuan Luo, and Isaac Kohane.
\newblock 2008.
\newblock Identifying patient smoking status from medical discharge records.
\newblock {\em Journal of the American Medical Informatics Association},
  15(1):14--24.

\bibitem[\protect\citename{Uzuner \bgroup et al.\egroup }2010]{uzuner10b}
{\"O}zlem Uzuner, Imre Solti, and Eithon Cadag.
\newblock 2010.
\newblock Extracting medication information from clinical text.
\newblock {\em Journal of the American Medical Informatics Association},
  17(5):514--518.

\bibitem[\protect\citename{Uzuner \bgroup et al.\egroup }2011]{uzuner10a}
{\"O}zlem Uzuner, Brett~R South, Shuying Shen, and Scott~L DuVall.
\newblock 2011.
\newblock 2010 i2b2/va challenge on concepts, assertions, and relations in
  clinical text.
\newblock {\em Journal of the American Medical Informatics Association},
  18(5):552--556.

\bibitem[\protect\citename{van Mulligen \bgroup et al.\egroup }2012]{van2012}
Erik~M. van Mulligen, Annie Fourrier-Reglat, David Gurwitz, Mariam Molokhia,
  Ainhoa Nieto, Gianluca Trifiro, Jan Kors, and Laura~I Furlong.
\newblock 2012.
\newblock {T}he {EU}-{ADR} corpus: {A}nnotated drugs, diseases, targets, and
  their relationships.
\newblock {\em Journal of biomedical informatics}, 45(5):879--884.

\bibitem[\protect\citename{Xu}2008]{xu2008}
Fei-Yu Xu.
\newblock 2008.
\newblock {\em Bootstrapping Relation Extraction from Semantic Seeds}.
\newblock {Ph.D.} thesis, Saarland University.

\bibitem[\protect\citename{Yin and Schütze}2015]{YinS15}
Wenpeng Yin and Hinrich Schütze.
\newblock 2015.
\newblock Multichannel variable-size convolution for sentence classification.
\newblock In Afra Alishahi and Alessandro Moschitti, editors, {\em CoNLL},
  pages 204--214. ACL.

\bibitem[\protect\citename{Zelenko \bgroup et al.\egroup }2003]{zelenko2003}
Dmitry Zelenko, Chinatsu Aone, and Anthony Richardella.
\newblock 2003.
\newblock Kernel methods for relation extraction.
\newblock {\em The Journal of Machine Learning Research}, 3:1083--1106.

\bibitem[\protect\citename{Zeng \bgroup et al.\egroup }2014]{Zeng14}
Daojian Zeng, Kang Liu, Siwei Lai, Guangyou Zhou, and Jun Zhao.
\newblock 2014.
\newblock Relation classification via convolutional deep neural network.
\newblock In {\em {COLING} 2014, 25th International Conference on Computational
  Linguistics, Proceedings of the Conference: Technical Papers, August 23-29,
  2014, Dublin, Ireland}, pages 2335--2344.

\end{thebibliography}
\bibliographystyle{acl2016}

\end{document}